%% file: paper.tex
\documentclass[runningheads]{llncs}

\usepackage{xcolor}         
\usepackage{amssymb,amsmath,enumitem}

\input{common/packages}

\begin{document}

\setlength{\abovedisplayskip}{6pt}
\setlength{\belowdisplayskip}{6pt}

\input{content/title}
\input{content/abstract}

\input{content/sections/1-introduction}
\input{content/sections/2-related_works}
\input{content/sections/3-method}
\input{content/sections/4-experiments}
\input{content/sections/5-conclusions}

\bibliographystyle{splncs04}
\bibliography{bibliography/references}

\end{document}

%% file: common/packages.tex
\usepackage{graphicx}

\usepackage{hyperref}

\usepackage{adjustbox}

\usepackage{multirow}

\usepackage{bm}

\usepackage{caption}

\usepackage{subcaption}

\usepackage{tabularray}

\usepackage{graphbox}

\usepackage{comment}

\usepackage{float}

\usepackage{caption}

\usepackage{amsmath}


%% file: content/title.tex
\title{
    Large Class Separation is not what you need for Relational Reasoning-based OOD Detection
}
\titlerunning{Class Separation in Relational Reasoning-based  OOD Detection}
%
\author{
    Lorenzo Li Lu\inst{1}\orcidID{0009-0001-9718-9422} \and
    Giulia D'Ascenzi\inst{1}\orcidID{0009-0003-3238-0300} \and
    Francesco Cappio Borlino\inst{1,2}\orcidID{0000-0002-8507-0213} \and \\ 
    Tatiana Tommasi\inst{1,2}\orcidID{0000-0001-8229-7159}
}

\authorrunning{Lu et al.}

\institute{Politecnico di Torino, Corso Duca degli Abruzzi 24, 10129 Torino, Italy \and Italian Institute of Technology, Italy \\
\email{\{lorenzo.lu, giulia.dascenzi\}@studenti.polito.it, \\ \{francesco.cappio, tatiana.tommasi\}@polito.it}\\}

\maketitle

%% file: content/abstract.tex
\begin{abstract}
Standard recognition approaches are unable to deal with novel categories at test time. Their overconfidence on the known classes makes the predictions unreliable for safety-critical applications such as healthcare or autonomous driving. Out-Of-Distribution (OOD) detection methods provide a solution by identifying semantic novelty. Most of these methods leverage a learning stage on the known data, which means training (or fine-tuning) a model to capture the concept of \emph{normality}. This process is clearly sensitive to the amount of available samples and might be computationally expensive for on-board systems. A viable alternative is that of evaluating similarities in the embedding space produced by large pre-trained models without any further learning effort. We focus exactly on such a fine-tuning-free OOD detection setting.

This works presents an in-depth analysis of the recently introduced relational reasoning pre-training and investigates the properties of the learned embedding, highlighting the existence of a correlation between the inter-class feature distance and the OOD detection accuracy.
As the class separation depends on the chosen pre-training objective, we propose an alternative loss function to control the inter-class margin, and we show its advantage with thorough experiments. 

\keywords{
    Out-Of-Distribution Detection
    \and Cross-Domain Learning
    \and Relational Reasoning
}
\end{abstract}

%% file: content/sections/1-introduction.tex
\section{Introduction}

In recent years, Deep Neural Networks have seen widespread adoption in multiple computer vision tasks. Still, standard recognition algorithms are typically evaluated under the \textit{closed-set} assumption \cite{ood_survey}, limiting their prediction
ability to the same categories experienced at training time. 
As most real-world scenarios are very different from the well-defined and controlled laboratory environments, an agent operating in the wild will inevitably face data coming from unknown distributions, thus it should be able to handle novelty which is a task of utmost importance for    
safety-critical applications. 
In this regard, Out-Of-Distribution (OOD) detection techniques have gained considerable attention as they enable models to recognize when test samples are \textit{In-Distribution} (ID) with respect to the training ones, or conversely \textit{Out-Of-Distribution} (OOD). 
Specifically, \emph{Semantic Novelty Detection} \cite{ood_survey} refers to the \emph{open-set} case in which the distribution shift originates from the presence of unknown categories in the test set, together with the known \emph{normal} ones already seen during training.
Many techniques have been proposed for this task \cite{msp,odin,energy_ood,react}. 
However, they typically need a significant number of reference known samples for the model to learn the concept of \textit{normality} through either training from scratch or at least a fine-tuning phase. While such approaches generally lead to good performance, they can also be problematic for low-power edge devices with limited computational resources, and anyway become unfeasible if the amount of known data is scarce or their training access is restricted for privacy reasons. 

\input{images/teaser}

Recently, two studies proposed techniques that enable OOD detection without fine-tuning \cite{resend,mcm}. Both rely on pre-trained models whose data representation can be easily exploited to perform comparisons and identify unknown categories, avoiding further learning effort. In terms of training objectives, they share the choice of moving away from standard classification and highlight the importance of analogy-based learning to better manage open-set conditions. Specifically, ReSeND \cite{resend} proposed a new relational reasoning paradigm to learn whether pairs of images belong or not to the same class. Instead, MCM \cite{mcm} inherits the CLIP model trained on vision-language data pairs to promote multi-modal feature alignment via contrastive learning.

In this work, we are interested in the single modality case to further evaluate its potential and limits. More precisely, we examine the connection between inter-class distance in the features embedding produced via relational reasoning and the ability to perform semantic novelty detection in that space. 
As highlighted in \cite{loss_transfer}, training objectives that enforce a stronger inter-class separation may cause the learned representations to be less transferable.
Thus, we present an extensive analysis of relational reasoning performed with various loss functions that have different control on class separation. Our findings indicate that avoiding to maximize inter-class separations provides more generalizable features, improving the performance of the pre-trained model on the downstream semantic novelty detection task (see Fig. \ref{fig:teaser}). Building on this conclusion, we design a tailored hinge loss function that provides direct control of class separation and increases the OOD results of the relational reasoning-based model.

Finally, we observe that certain OOD detection methods based on classification pre-training and originally intended to be used via fine-tuning may skip the latter learning phase \cite{mahalanobis,knn_ood}. Hence, these approaches can serve as a fair benchmark reference for relational reasoning-based methods. 

\smallskip
\noindent To summarize, our key contributions are:
\begin{itemize}[topsep=0pt]
    \item We discuss and evaluate how the feature distributions originating from the use of different pre-training objectives affect the capability of a relational reasoning model for OOD detection; 
    \item We introduce an alternative loss function that provides better control of class-specific feature distributions;
    \item  We run a thorough experimental analysis that demonstrates the advantages of the proposed loss,  considering as a reference also the  powerful but costly \textit{k}-NN-based OOD detector \cite{knn_ood}, re-casted for the first time to work in the fine-tuning-free setting.
\end{itemize}

%% file: images/teaser.tex
\begin{figure}[t]
    \centering
    \includegraphics[width=0.9\textwidth]{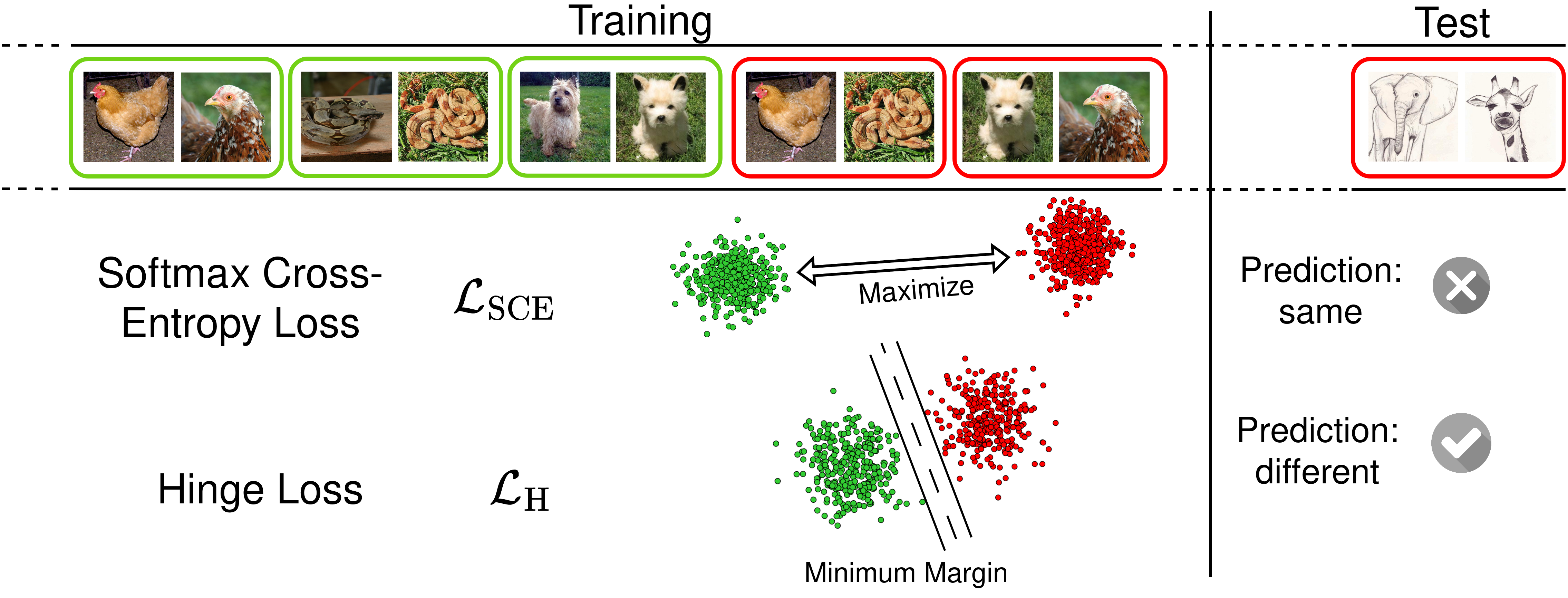}
    \caption{Schematic overview of a relational reasoning-based OOD method that exploits different training losses. Our work empirically demonstrates that controlling, and in particular reducing, 
    the distance between \emph{same} and \emph{different} classes improves semantic novelty detection.}\vspace{-5mm}
    \label{fig:teaser}
\end{figure}

%% file: content/sections/2-related_works.tex
\section{Related Work}
\noindent\textit{Out-Of-Distribution detection} is the task of determining whether test data belong to the same distribution as the training data or not. 
A distributional shift may occur due to a change in domain (\textit{covariate shift}) or categories (\textit{semantic shift}). 
We expect a trustworthy model to detect whether a sample belongs to a new category regardless of the visual domain, thus our main focus is on semantic novelty detection. 
The first baseline for this task was proposed by Hendrycks et al. \cite{msp}, who suggested that the Maximum Softmax Probability (MSP) score produced by a classification model trained on ID data should be higher for ID test samples than for OOD ones. Several other approaches followed the same post-hoc strategy enhancing ID-OOD separation, via temperature scaling \cite{odin}, or by focusing on energy scores \cite{energy_ood} and network unit activations \cite{react}.
A different family of techniques uses distance metrics to identify OOD samples in the feature space learned on the ID data \cite{mahalanobis,knn_ood}. 

\noindent\textit{OOD detection without fine-tuning.}
All the OOD detection solutions described in the previous paragraph require training (or at least fine-tuning) on nominal samples in order to learn the concept of normality. However, this learning phase requires a sizable amount of ID data and computational resources, making it expensive and impractical for many real-world applications. Additionally, fine-tuning can hurt the generalization of learned representations \cite{kumar_finetuning_ICLR2022} as it is susceptible to \emph{catastrophic forgetting} \cite{forgetting}: the model may overfit the fine-tuning dataset, losing the knowledge previously learned on a much larger one.  Only some recent work has started addressing this problem, proposing solutions that do not require a fine-tuning stage to perform OOD detection \cite{resend,mcm}.  
In particular, \cite{resend} suggests substituting the standard classification-based pre-training task with relational reasoning, which directs the network's focus on the semantic similarity between two input images to predict a normality score. This pretext objective is less domain- and task-dependent than classification and leads to an embedding space with great transfer capabilities. On the other hand, \cite{mcm} leverages CLIP \cite{clip} to perform zero-shot OOD detection, thus exploiting two modalities (vision and language) rather than one. Finally, we point out how distance-based strategies such as \cite{mahalanobis,knn_ood} could also be used without performing the fine-tuning stage, although this aspect was not addressed in their original works. 

\noindent\textit{Pre-training loss functions and transfer learning.}
The possibility to easily inherit and reuse pre-trained models for novel tasks is certainly one of the more appreciated characteristics of deep learning. As discussed above, such a procedure may be relevant even for OOD detection applications in which the representation learned on a large-scale dataset is leveraged to evaluate sample similarity. 
Still, only a few works have analyzed how the exact choice of the pre-training objective influences the transferability of the extracted knowledge. An implicit hypothesis is that models that perform well on the pre-training task also perform well on the downstream one. 
However, this is not always the case \cite{imagenet_transfer}: for instance, some regularization techniques that provide an improvement on the pre-training task produce penultimate layers features that are worse in generalization. This phenomenon has been described as \textit{supervision collapse} \cite{supervision_collapse}. The $R^2$ metric introduced in \cite{loss_transfer} to evaluate intra-class compactness and inter-class separation provides a way to shed light on this behavior: the most advanced strategies to increase accuracy on the pre-training task lead to a greater class separation which however is associated with reduced knowledge transferability.
As the use of pre-trained models without fine-tuning for OOD detection is still scarcely explored, we find it relevant to perform an analysis of the role of different pre-training objectives for this downstream task. Specifically, we focus on relational reasoning-based OOD detection performed via different loss functions.

%% file: content/sections/3-method.tex
\section{Relational reasoning for OOD detection}

We consider a set of labeled samples $\mathcal{S}=\{\bm{x}^s, y^s\}$ that we call \textit{support set}, and a set of unlabeled ones $\mathcal{T}=\{\bm{x}^t\}$ called \textit{test set}. They are drawn from two different distributions and present a category shift, besides also a potential domain shift. The support label set $\mathcal{Y}_{\mathcal{S}}$ identifies \textit{known} categories. The target label set $\mathcal{Y}_{\mathcal{T}}$ includes both known and unknown semantic categories: $\mathcal{Y}_{\mathcal{S}} \subset \mathcal{Y}_{\mathcal{T}}$. The goal of an OOD detector is to identify all the test samples whose categories do not appear in the support set (i.e., which are \textit{unknown}). Traditional methods require a training or fine-tuning stage on the support set $\mathcal{S}$, while in the fine-tuning-free scenario the support set is only accessed at evaluation time. 

In ReSeND \cite{resend}, the authors presented a relational reasoning-based learning approach specifically designed for OOD detection. The model is trained on sample pairs $(\bm{x}_i, \bm{x}_j)$ drawn from a large-scale object recognition dataset and learns to distinguish whether the two images belong to the same category ($l_{ij}=1$, if $y_i=y_j$) or not ($l_{ij}=0$, if $y_i\neq y_j$). This task can be cast as binary classification or regression. In both cases, the model learns how to encode in an embedding space the samples' semantic relationship $\bm{p}_m=r(\bm{z}_i,\bm{z}_j)$, where the index $m$ ranges over all the possible sample pairs, and $\bm{z}=\phi(\bm{x})$ represents the features extracted via an encoder $\phi$ from the image $\bm{x}$. Then, the last network layer converts this information into a scalar similarity value that is compared to the ground truth $l_m$ with a chosen loss function.
At inference time, the support set samples are grouped according to their category and their representation is averaged to get per-class prototypes $\bar{\bm{z}}^s_y$ for $y=\{1, \ldots,|\mathcal{Y}_{\mathcal{S}}|\}$.
Each test sample $\bm{z}^t=\phi(\bm{x}^t)$ is then compared with every prototype to get the corresponding similarity score. Finally, the vector collecting all the $|\mathcal{Y}_{\mathcal{S}}|$ elements is filtered by a softmax function on which MSP is applied to get the final normality score.

In this framework, by observing the embedding space produced by the penultimate layer of the network, we expect to see pairs of samples of the training dataset  organized into two clusters representing the broad \emph{same} and \emph{different} concept classes. Once trained on the large-scale ImageNet-1K dataset, this embedding can be used for OOD detection on a variety of domains without fine-tuning, so its generalization ability is crucial.

\section{Relational reasoning and class separation}
\subsection{Class compactness and separation}
\label{ssec:r2}
In order to analyze the learned feature space we focus on the separation between the \textit{same} and \textit{different} classes described above. 
In particular, we leverage the $R^2$ index introduced in \cite{loss_transfer}. This metric is based on the ratio between the average within-class and average global cosine distance for the considered feature vectors, providing a relative measure of the sparsity of the representation of each class in the embedding space. Specifically, the index value is given by:
\begin{equation} \label{eq:r2}\vspace{2mm}
    R^2 = 1 - \bar{d}_{within} / \bar{d}_{total}
\end{equation}

{\small \noindent
$\bar{d}_{within} = \sum_{k=1}^K \sum_{i=1}^{M_k} \sum_{j=1}^{M_k}
    \frac{1 - \textrm{sim}(\bm{p}^k_i,\bm{p}^k_j)}{K M_k^2}$, $  \bar{d}_{total} = \sum_{h=1}^{K} \sum_{k=1}^{K} \sum_{i=1}^{M_h} \sum_{j=1}^{M_k}
    \frac{1 - \textrm{sim}(\bm{p}^h_i,\bm{p}^k_j)}{K^2 M_h M_k}$}

\noindent where the indices $i,j \in \{1, \ldots, M_k\}$ now range on the pairs of samples $\bm{p}^k$ which belong respectively to the $K=2$ classes. The relative distance is measured via the cosine similarity: $\textrm{sim}(\bm{a},\bm{b}) = \bm{a}^T \bm{b} / (\Vert \bm{a} \Vert \Vert \bm{b} \Vert)$. The right part of Fig. \ref{fig:scatter} gives an idea of what high and low $R^2$ values mean in terms of class separation.

\subsection{Relational Reasoning Loss Functions}
\label{ssec:losses}
In the following we review some of the most common loss functions used for binary problems. 
In all the loss equations we use $\sigma$ to refer to the score produced as output by the network for a sample pair $\bm{p}$, while the ground truth label is $l$. 

\smallskip
\noindent \textbf{Binary Cross-Entropy.}
The Cross-Entropy loss is defined as:
\begin{equation} \label{eq:ce}
    \mathcal{L}_{\textrm{CE}}=-\sum_{m=1}^M \sum_{k=1}^{K} t_{m,k} \log (\hat{t}_{m,k})
\end{equation}
where $K$ is the number of categories, while $t_{m,k}$ and $\hat{t}_{m,k}$ are respectively the target value and the predicted probability of the class $k$ for the sample $m$. In particular $t_{m,k}$ will assume the value 1 for the ground truth class of the sample ($k = l_m$) and 0 for all the other categories (one-hot encoding). In the binary case (i.e., when $K = 2$), such loss function can be expressed as:

\begin{equation} \label{eq:bce}
    \mathcal{L}_{\textrm{BCE}} = - \sum_{m=1}^{M} \left(t_{m,1}\log(1-\hat{t}_{m,2}) + t_{m,2}\log(\hat{t}_{m,2})\right)
\end{equation}
where $\hat{t}_{m,2}$ is generally obtained by applying the logistic sigmoid function to the model output score ($f(\sigma)=1/(1+e^{-\sigma})$).  

\noindent \textit{Impact on the class separation}: this loss is non-zero even for correctly classified samples. As a result, the intra-class compactness and inter-class separation keep increasing for the whole training procedure.

\smallskip
\noindent \textbf{Softmax Cross-Entropy.}
The categorical Cross-Entropy loss that is generally adopted for multi-class problems, is obtained by using the Cross-Entropy in Eq. (\ref{eq:ce}) after having applied the softmax function to the model output scores ($f(\sigma)_k=e^{\sigma_k}/\sum_{c=1}^C e^{\sigma_c}$). 
Considering that the labels are one-hot, the overall summation will contain for each sample only the term corresponding to its ground truth label, so we can write the Softmax Cross-Entropy as:
\begin{equation} \label{eq:sce}
    \mathcal{L}_{\textrm{SCE}} = - \sum_{m=1}^{M} \log \frac{e^{\sigma_{m,l_m}}}{\sum_{k=1}^{K} e^{\sigma_{m,k}}}
\end{equation}
where $\sigma_{m,k}$ is the score corresponding to the class $k$ for the sample $m$ and $l_m$ represents its
ground truth label. In the binary case we suppose $k, l \in \{1,2\}$.

\noindent \textit{Impact on the class separation}:
As in the previous BCE case, this loss is non-zero even for correctly classified samples. It has been shown that the consequent trend of growing intra-class compactness and inter-class separation leads to miscalibrated classifiers providing overconfident predictions \cite{logit_norm,calibrating_focal}.

\smallskip
\noindent \textbf{Focal Loss.}
A possible solution for the miscalibration issue mentioned above is to adjust the penalty assigned to a sample based on the network's confidence in predicting its true class  \cite{calibrating_focal}.
This can be accomplished with the Focal Loss \cite{focal}.
Starting from the Cross-Entropy formulation (see Eq. \ref{eq:ce}), such loss function can be expressed as:
\begin{equation} \label{eq:focal}
    \mathcal{L}_{\textrm{focal}} = - \sum_{m=1}^{M} \sum_{k=1}^{K} t_{m,k} (1 - \hat{t}_{m,k})^{\gamma} \log(\hat{t}_{m,k})
\end{equation}
where $\gamma$ is a hyperparameter controlling the rescaling strength.

\noindent \textit{Impact on the class separation}:
by varying $\gamma$, it's possible to tune the magnitude of the rescaling, effectively bringing the loss
value for correctly classified samples near zero and therefore mitigating the class separation tendency.

\smallskip
\noindent \textbf{MSE with a compressed sigmoid.}
\input{images/fn_visualizations}
In ReSeND \cite{resend} the problem of separating the same and different classes was formalized as a regression task by using the MSE loss computed between the ground truth $l_m \in \{-1, 1\}$ and the output provided by a sigmoid rescaled on the $[-1, 1]$ range and with a modified slope, controlled by a factor $c$ (see Fig. \ref{fig:sigmoid_compression} (a)): 
\begin{equation} \label{eq:mse}
    \mathcal{L}_{MSE} = \sum_{m=1}^{M} (\hat{s}_c(\sigma_m) - l_m)^2 \qquad \textrm{with} \quad \hat{s}_c(\sigma_m) = \frac{2}{1 + e^{-c\sigma_m}} - 1
\end{equation}

\noindent \textit{Impact on the class separation}:
by varying the value of $c$, it is possible to tune the penalty associated with different scores $\sigma$. Specifically, for higher $c$ values, the sigmoid function will be more horizontally compressed: as a consequence samples already correctly classified receive a loss value that is almost zero decreasing the need for further class separation. 

\subsection{Controlling class separation: Hinge Loss for relational reasoning}

As it is clear that class separation is crucial for the problem, we introduce a loss function that allows us to precisely tune it in a simple and straightforward way.

Let's start from the output of the last layer which is a scalar score $\sigma_m$ and can be positive or negative, indicating the corresponding two classes. We can simply set a threshold at zero and fix a margin $\delta$ around it, within which even correct predictions pay a penalty. The loss will cancel out for $\sigma_m >\delta$ on positive samples and $\sigma_m <-\delta$ for negative ones, but would grow linearly if a negative score is assigned to a positive sample and vice-versa (see Fig. \ref{fig:sigmoid_compression} (b)). 

In this way we keep the two classes separated (which is crucial to retain the model's discriminative power), but the margin is limited and fixed to $\delta$. This formulation corresponds to a hinge loss applied on the scalar score $\sigma_m$: 
\begin{equation} \label{eq:margin_loss}
    \mathcal{L}_{H} = \sum_{m=1}^{M} \max(0, \delta - l_m \sigma_m) \qquad \textrm{with} \quad l_m \in \{-1, 1\}
\end{equation}

%% file: images/fn_visualizations.tex
\begin{figure}[t]
    \centering
    \hspace{25pt}
    \begin{subfigure}{.30\textwidth}
        \centering
        \includegraphics[width=\textwidth]{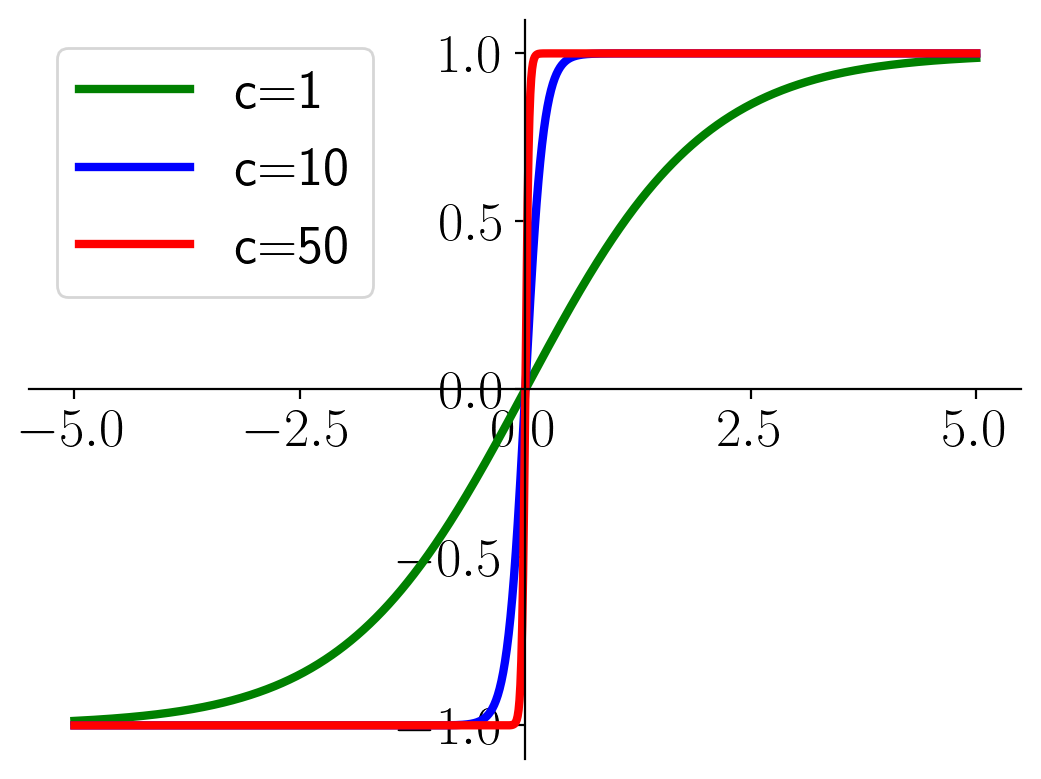}
        \caption{Compressed sigmoid}
        \label{fig:sigmoid_comp}
    \end{subfigure}%
    \hfill
    \begin{subfigure}{.30\textwidth}
        \centering
        \includegraphics[width=\textwidth]{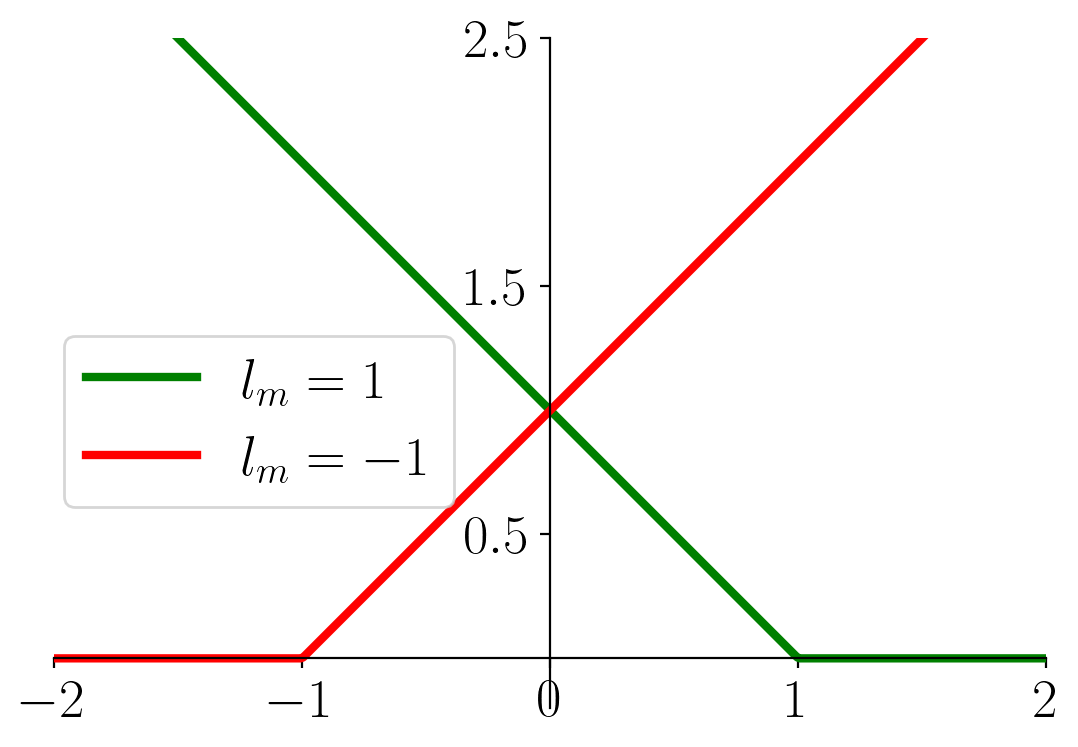}
        \caption{Hinge loss trend}
    \end{subfigure}%
    \hspace{25pt}
    \caption{(a) Increasing $c$ in the MSE compressed sigmoid transforms it into a Heaviside step function: the loss is zero when the output score has the correct sign. (b) H loss trend for positive ($l_m=1$) and negative ($l_m=-1$) pairs ($\delta=1$).} \vspace{-5mm}
    \label{fig:sigmoid_compression}
\end{figure}

%% file: content/sections/4-experiments.tex
\section{Experiments}

\subsection{Experimental protocol}
Our experimental analysis presents an extensive benchmark of fine-tuning-free OOD methods. All of them consist of a pre-training phase on ImageNet-1K \cite{imagenet} with a different objective, followed by a distance-based OOD prediction protocol. For ReSeND \cite{resend} the pre-training task is relational reasoning (same vs different) executed with all the loss functions described in the previous Section. The other competitors exploit either supervised classification or self-supervised objectives, with both cross-entropy-based approaches (ResNet\cite{resnet}, ViT\cite{vit}, CutMix\cite{cutmix}), and contrastive strategies (SimCLR \cite{simclr}, SupCLR \cite{supclr}, CSI \cite{csi}, SupCSI \cite{csi}). We also evaluate Mahalanobis \cite{mahalanobis} and \textit{k}-NN \cite{knn_ood}. We emphasize that the \textit{k}-NN approach has never been previously evaluated in a fine-tuning-free setting. We include it in our comparison despite its potentially higher computational cost, as it involves comparing the test sample with each support set instance (which must be stored in memory), rather than with a single prototype per class.

Unless otherwise specified, we always adopt a ResNet-101 backbone, as it includes a comparable number of parameters to ReSeND (44M and 40M, respectively).
We publish the code, together with implementation details and additional results in our project page \footnote{\href{https://github.com/lulor/ood-class-separation}{https://github.com/lulor/ood-class-separation}}.

We adopt two different experimental set-ups, by following \cite{resend}. 
The \emph{intra-domain} setting is designed to evaluate the OOD detection ability of a model when there is a purely semantic distribution shift between the support and the test sets. It is built upon the DomainNet \cite{domainnet} and DTD \cite{dtd} datasets.
In the \emph{cross-domain} setting the support and test set are sampled from different domains so we can evaluate the ability of the OOD methods to focus on semantics and disregard other visual appearance discrepancies. Rather than using the limited PACS dataset \cite{li_2017_pacs} as done in \cite{resend}, we propose a novel benchmark built on top of DomainNet \cite{domainnet}. This choice allows for more statistically significant results. 

Following common practice, we report results in terms of Area Under the Receiver Operating Characteristic curve (\textit{AUC}) and FPR@TPR95 (\textit{FPR}), which indicates the false positive rate value when the ID true positive rate is 95\%. 

\input{images/loss_comparison}
\begin{figure}[t]
    \centering

    \begin{subfigure}{.33\textwidth}
        \centering
        \includegraphics[width=\textwidth]{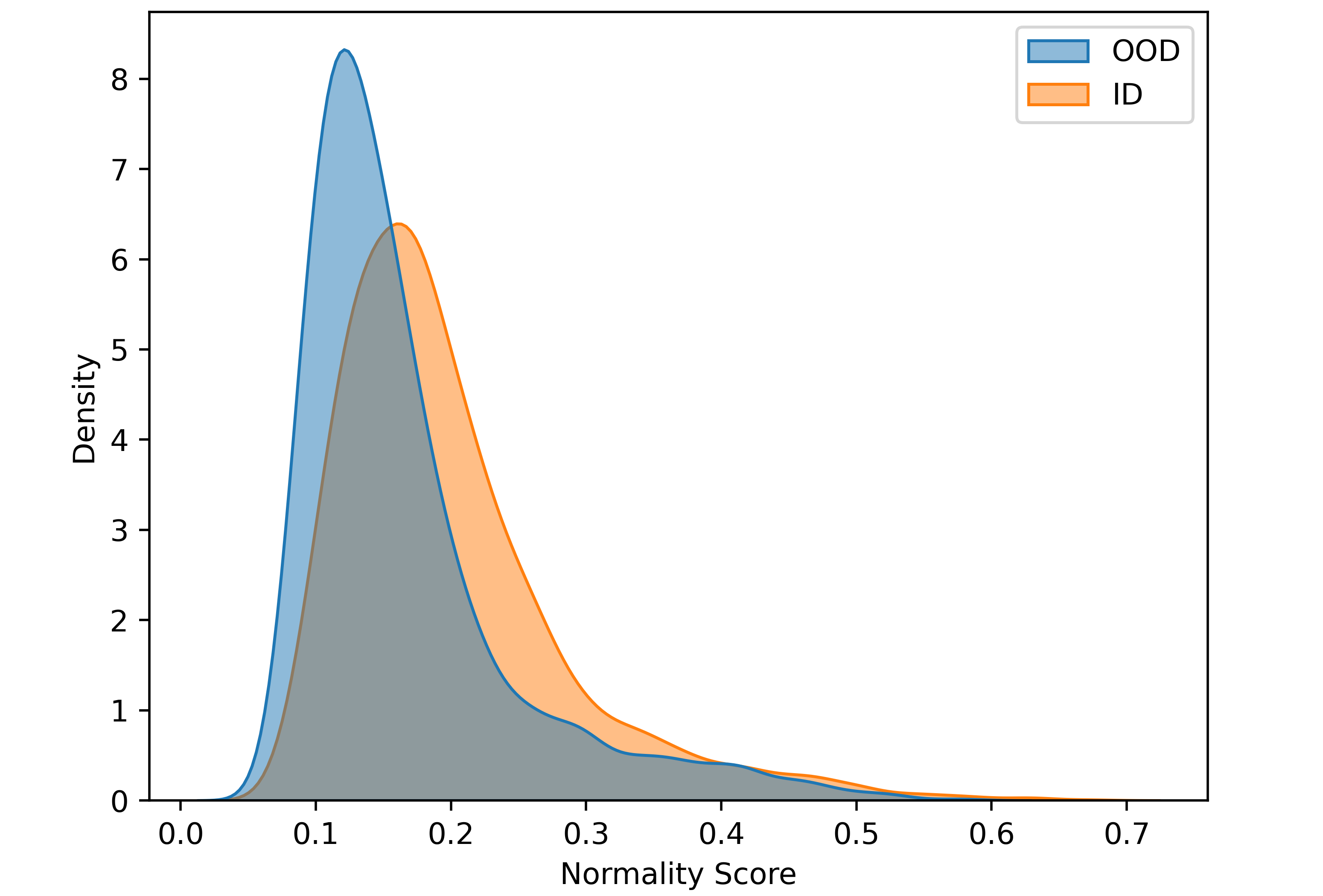}
        \caption{SCE}
    \end{subfigure}%
    \begin{subfigure}{.33\textwidth}
        \centering
        \includegraphics[width=\textwidth]{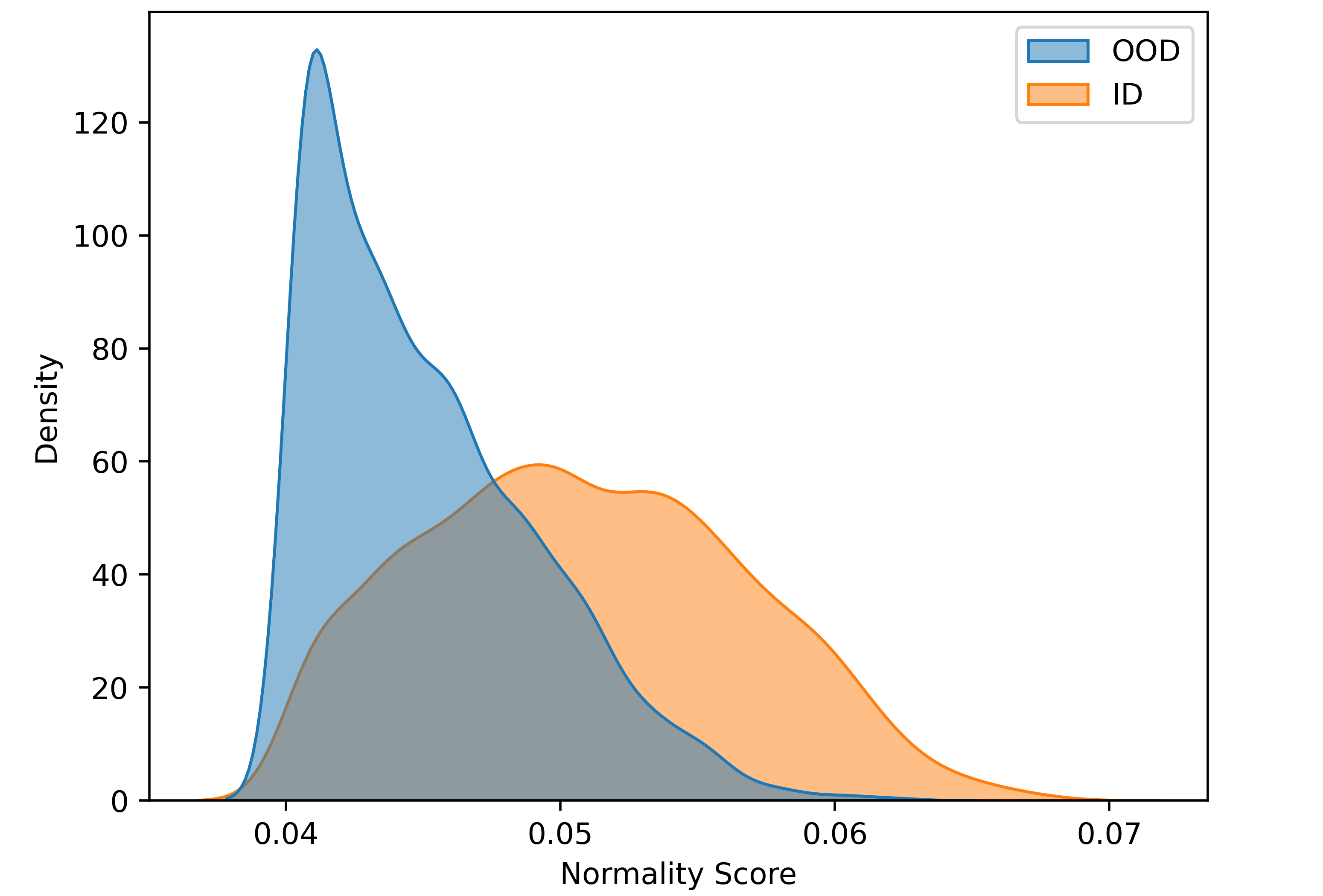}
        \caption{H, $\delta$=0.1}
    \end{subfigure}%
    \begin{subfigure}{.33\textwidth}
        \centering
        \includegraphics[width=\textwidth]{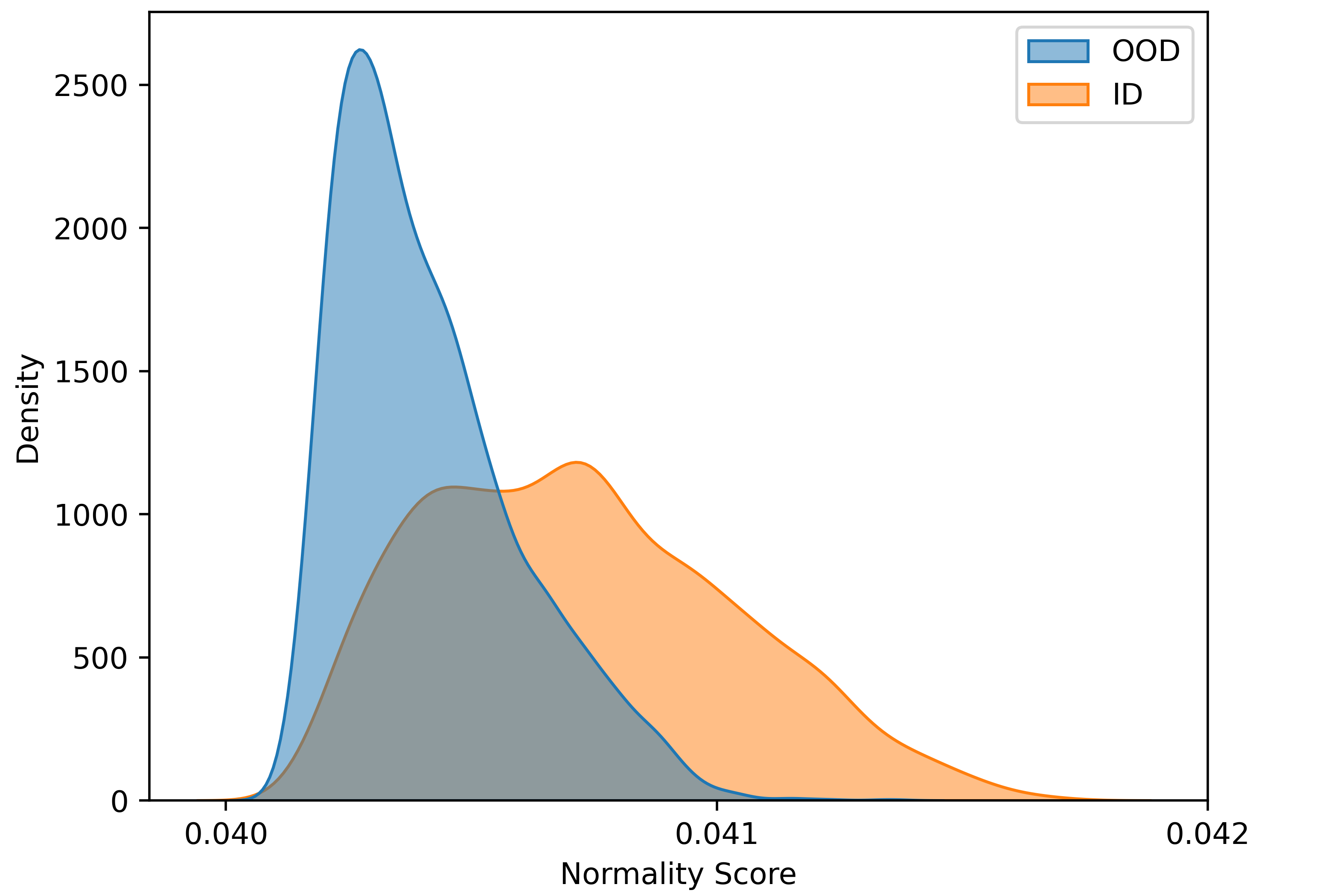}
        \caption{H, $\delta$=0.01}
    \end{subfigure}

    \caption{
        Normality Score distributions on the intra-domain Real setting for ReSeND pre-trained with different loss functions.
        We can see how the hinge loss with low margin pushes the model to provide more conservative scores, which are very close to each other (check the horizontal axis' scale) but more discernible.
    }
    \label{fig:msp}\vspace{-3mm}
\end{figure}

\subsection{Impact of the training objective}
We evaluate the impact on ReSeND of various loss functions. As the learning objective shapes the structure of the feature space, we can investigate how the distribution of the data in the learned embedding relates to the final OOD performance. For this analysis we focus on the intra-domain setting. The average AUC on the four datasets, along with the corresponding $R^2$ value, are reported in the scatter plot in the left part of Fig. \ref{fig:scatter}. Detailed per-dataset results can be found on our project page.
The results clearly highlight a general trend in which a higher inter-class separation is associated with a lower OOD detection performance. This behavior is even more evident when focusing on a specific loss and looking at how the results change by varying its hyperparameter value (e.g. when changing $\delta$ for our H loss or $c$ for the MSE).
Of course, there is a limit in the performance gain that can be reached by reducing the inter-class separation: after a certain point, the features start losing their discriminative power. 
For example with $c \geq 50$ for the MSE case, the training becomes less effective  and the OOD detection performance starts slowly decreasing. 
We can conclude that for relational reasoning it is important to choose a learning objective that allows for a precise margin control. 
Only the proposed $\mathcal{L}_{H}$ loss satisfies this condition. Its hyperparameter $\delta$ represents a geometrical margin, and when a training sample meets the margin condition, the sample loss no longer affects the learning process. This behavior avoids the overconfidence typical of the standard softmax cross-entropy as highlighted by the normality score distributions represented in Fig. \ref{fig:msp}: the normality score values provided by the SCE loss are generally higher and have a larger range than those provided by the $\mathcal{L}_{H}$ loss (see the horizontal axis), but at the same time they provide a weaker ID-OOD separation.

\subsection{Intra-Domain and Cross-Domain OOD Results}
\noindent\textit{Intra-Domain analysis.}
\input{content/tables/baselines_intra}
In this setting support and test sets only differ in terms of semantics. Still, with respect to the pre-training dataset (ImageNet-1K), there may be a domain shift of varying magnitude (smaller for the Real case, larger for the others). 
In Table \ref{tab:baselines_intra} we collect the results of the original ReSeND formulation ($\mathcal{L}_{MSE}$, with $c=10$), its version based on the hinge loss that we name ReSeND-H ($\mathcal{L}_{H}$, with $\delta=0.01$) as well as the ResNet baseline and several reference approaches. We observe that 
ReSeND-H obtains a small but meaningful improvement across most of
the considered settings, particularly in the Real and Sketch ones.
The \textit{k}-NN method from \cite{knn_ood}, applied without fine-tuning,
achieves the best performance among all the considered techniques. 
We highlight how this result comes with a significant cost in terms of memory usage. Specifically, we calculated the average number of comparisons (n.comp) per test sample needed at evaluation time and reported the corresponding value in the last table column. 
Indeed, this introduces an important scalability limitation. 

\smallskip
\noindent\textit{Cross-Domain analysis.}
\input{content/tables/baselines_cross}
In this setting train and test data differ in semantic content and in visual style. From the results in  Table \ref{tab:baselines_cross} we can see how, despite using the whole support set rather than just the class prototypes, the \textit{k}-NN method does not achieve the same performance advantage exhibited in the intra-domain case. Indeed, relying on all the support samples appears misleading. As a consequence, this method is less robust to domain shifts compared to ReSeND and ReSeND-H, which instead shows similar performance to the corresponding one in the intra-domain setting. 

%% file: images/loss_comparison.tex
\begin{figure}[t]
    \centering
    
        \hspace*{\fill}%
        \begin{minipage}[t]{0.6\textwidth}
            \centering
            \begin{figure}[H]
                \centering
                \includegraphics[width=0.95\textwidth]{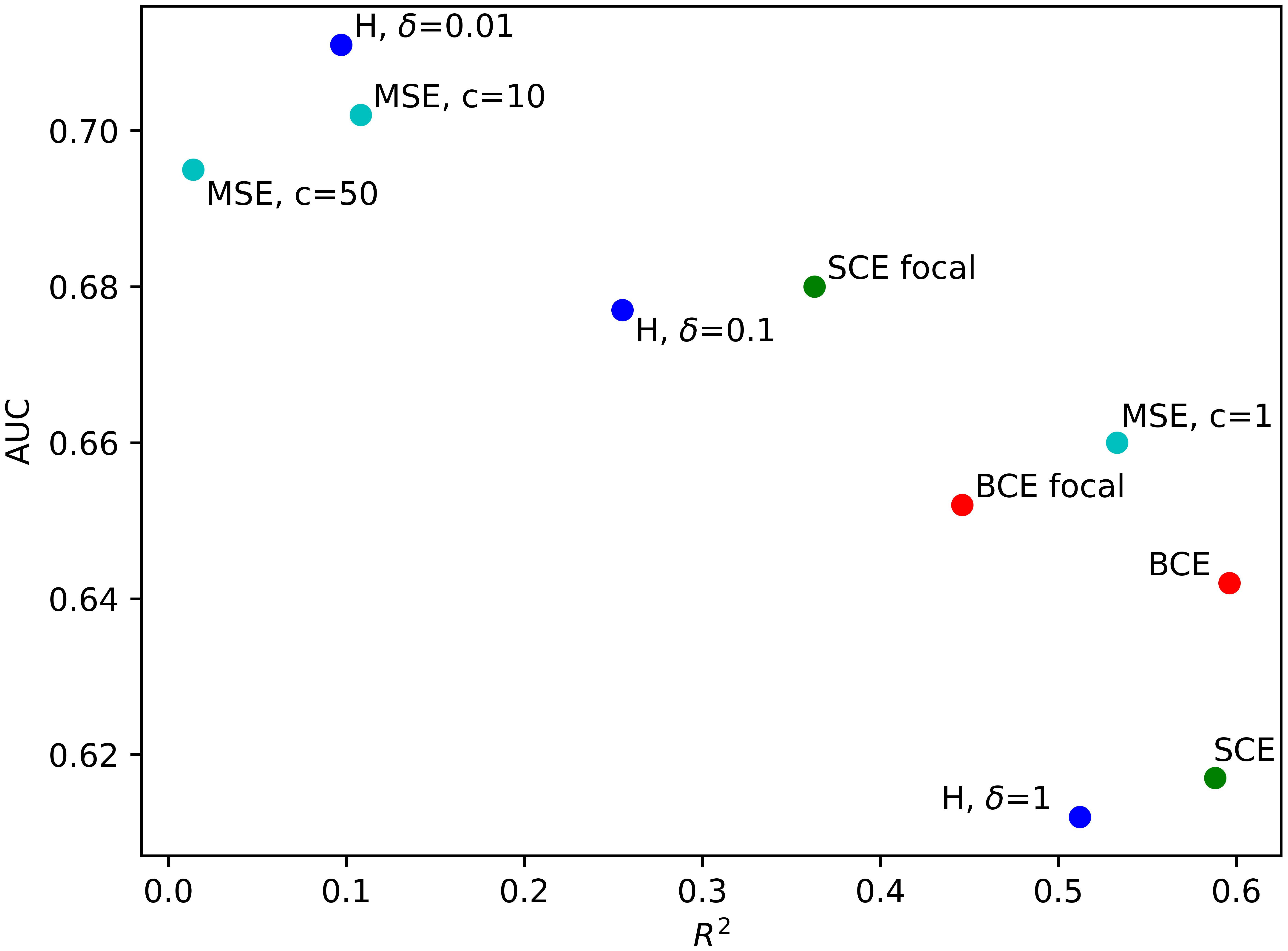}
                \label{fig:scatter}
            \end{figure}
        \end{minipage}%
        \hfill
        \begin{minipage}[t]{0.30\textwidth}
            \centering
            \begin{figure}[H]
                \centering
                \includegraphics[width=0.95\textwidth]{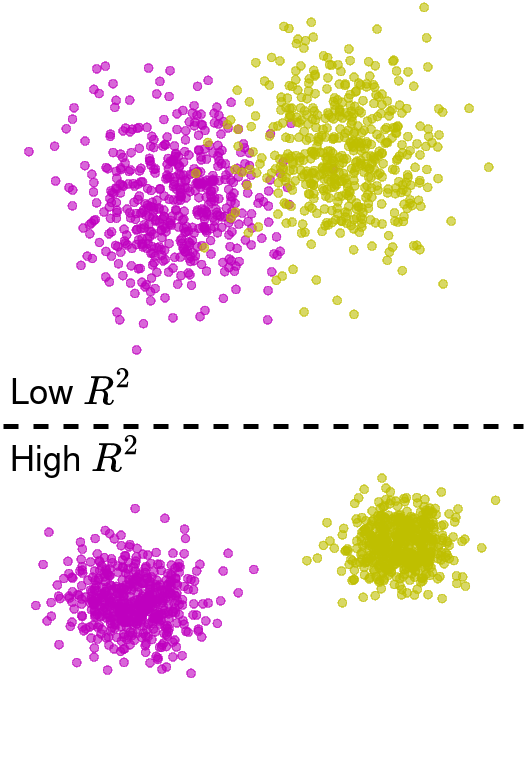}
                \label{fig:r2_qualitative}
            \end{figure}
        \end{minipage}%
        \hspace*{\fill}

    \caption{Analysis of the OOD performance of a relational reasoning-based model trained with different loss functions (average \textit{intra-domain} benchmarks results). The scatter-plot on the left shows that lower OOD results are generally associated with higher $R^2$ values, which means a stronger class separation as in the point distribution shown in the right-bottom part. On the other hand, more generalizable features and higher OOD performance are obtained with lower $R^2$ values corresponding to minimal class separation as in the right-top part.} \vspace{-5mm}
    \label{fig:scatter}
\end{figure}

%% file: content/tables/baselines_intra.tex
\begin{table}[t]
    \centering
    \caption[Intra-Domain results]{
        Intra-domain setting. 
        Best result in bold and second best underlined
    }
    \begin{adjustbox}{max width=\textwidth}
        \begin{tabular}{ |c|cc|cc|cc|cc|cc|c| }
            \hline
            \multirow{2}{*}{\textbf{Model}}
            & \multicolumn{2}{c|}{\textbf{Texture}}
            & \multicolumn{2}{c|}{\textbf{Real}}
            & \multicolumn{2}{c|}{\textbf{Sketch}}
            & \multicolumn{2}{c|}{\textbf{Painting}}
            & \multicolumn{2}{c|}{\textbf{Avg}}
            & \textbf{Avg}
            \\
            & AUC $\uparrow$ & FPR $\downarrow$
            & AUC $\uparrow$ & FPR $\downarrow$
            & AUC $\uparrow$ & FPR $\downarrow$
            & AUC $\uparrow$ & FPR $\downarrow$
            & AUC $\uparrow$ & FPR $\downarrow$
            & n.comp $\downarrow$
            \\
            \hline
            ResNet              & 0.672 & 0.897 & 0.710 & 0.863 & 0.554 & 0.939 & 0.649 & 0.919 & 0.646	& 0.904 & 25 \\
            ViT                 & 0.537 & 0.937 & 0.701 & 0.829 & 0.553 & 0.955 & 0.673 & 0.853 & 0.616 & 0.894 & 25 \\
            CutMix              & 0.605 & 0.925 & 0.722 & 0.876 & 0.544 & 0.944 & 0.627 & 0.929 & 0.625 & 0.919 & 25 \\
            SimCLR              & 0.526 & 0.942 & 0.475 & 0.943 & 0.489 & 0.955 & 0.508 & 0.959 & 0.500 & 0.950 & 25 \\
            SupCLR              & 0.588 & 0.921 & 0.496 & 0.956 & 0.481 & 0.953 & 0.514 & 0.959 & 0.520 & 0.947 & 25 \\
            CSI                 & 0.627 & 0.898 & 0.695 & 0.850 & 0.513 & 0.960 & 0.613 & 0.912 & 0.612 & 0.905 & 25 \\
            SupCSI              & 0.662 & 0.896 & 0.716 & 0.864 & 0.521 & 0.957 & 0.640 & 0.902 & 0.635 & 0.904 & 25 \\
            Mahalanobis         & 0.656 & 0.911 & 0.744 & 0.850 & 0.590 & 0.928 & 0.710 & 0.857 & 0.675 & 0.886 & 25 \\
            \hline
            ReSeND              & 0.684 & \underline{0.847} & 0.782 & 0.777 & 0.610 & 0.934 & \underline{0.721} & \underline{0.826} & 0.699	& 0.846 & 25 \\
            ReSeND-H           & \underline{0.688} & 0.885 & \underline{0.798} & \underline{0.755} & \underline{0.638} & \textbf{0.898} & 0.719 & \underline{0.826} & \underline{0.711} & \underline{0.841} & 25 \\
            \hline
            \textit{k}-NN (k=1)   & \textbf{0.774} & \textbf{0.840} & \textbf{0.843} & \textbf{0.596} & \textbf{0.640} & \underline{0.914} & \textbf{0.800} & \textbf{0.757} & \textbf{0.764} & \textbf{0.777} & 4100 \\
            \hline
        \end{tabular}
    \end{adjustbox}
    \label{tab:baselines_intra}\vspace{-3mm}
\end{table}

%% file: content/tables/baselines_cross.tex
\begin{table}[t]
    \centering
    \caption[Cross-Domain results]{Cross-domain setting. Best result in bold and second best underlined.
    }
    \begin{adjustbox}{width=1\textwidth}
        \begin{tabular}{ |c|cc|cc|cc|cc|cc|cc|cc|c| }
            \hline
            \multirow{2}{*}{\textbf{Model}}
            & \multicolumn{2}{c|}{\textbf{Real-Paint.}}
            & \multicolumn{2}{c|}{\textbf{Real-Sketch}}
            & \multicolumn{2}{c|}{\textbf{Paint.-Real}}
            & \multicolumn{2}{c|}{\textbf{Paint.-Sketch}}
            & \multicolumn{2}{c|}{\textbf{Sketch-Real}}
            & \multicolumn{2}{c|}{\textbf{Sketch-Paint.}}
            & \multicolumn{2}{c|}{\textbf{Avg}}
            & \textbf{Avg}
            \\
            & AUC $\uparrow$ & FPR $\downarrow$
            & AUC $\uparrow$ & FPR $\downarrow$
            & AUC $\uparrow$ & FPR $\downarrow$
            & AUC $\uparrow$ & FPR $\downarrow$
            & AUC $\uparrow$ & FPR $\downarrow$
            & AUC $\uparrow$ & FPR $\downarrow$
            & AUC $\uparrow$ & FPR $\downarrow$
            & n.comp $\downarrow$
            \\
            \hline
            ResNet                  & 0.596 & 0.949 & 0.539 & 0.938 & 0.627 & 0.922 & 0.546 & 0.941 & 0.533 & 0.929 & 0.524 & 0.940 & 0.561 & 0.937 & 25 \\
            ViT                     & 0.627 & 0.921 & 0.526 & \underline{0.931} & 0.618 & 0.901 & 0.524 & 0.946 & 0.568 & 0.945 & 0.591 & 0.924 & 0.576 & 0.928 & 25 \\
            CutMix                  & 0.585 & 0.940 & 0.533 & 0.944 & 0.630 & 0.915 & 0.534 & 0.949 & 0.550 & 0.939 & 0.530 & 0.950 & 0.560 & 0.940 & 25 \\
            SimCLR                  & 0.499 & 0.965 & 0.486 & 0.949 & 0.465 & 0.961 & 0.489 & 0.956 & 0.496 & 0.961 & 0.419 & 0.966 & 0.476 & 0.960 & 25 \\
            SupCLR                  & 0.507 & 0.966 & 0.471 & 0.959 & 0.468 & 0.962 & 0.469 & 0.957 & 0.524 & 0.968 & 0.463 & 0.965 & 0.484 & 0.963 & 25 
            \\
            CSI                     & 0.585 & 0.942 & 0.531 & 0.943 & 0.689 & \underline{0.863} & 0.503 & 0.953 & 0.552 & 0.867 & 0.448 & 0.942 & 0.551 & 0.918 & 25 \\
            SupCSI                  & 0.586 & 0.943 & 0.492 & 0.963 & 0.658 & 0.898 & 0.473 & 0.957 & 0.490 & 0.963 & 0.434 & 0.973 & 0.522 & 0.949 & 25 \\
            Mahalanobis             & 0.612 & 0.945 & 0.564 & 0.938 & 0.646 & 0.943 & 0.577 & 0.928 & 0.577 & 0.912 & 0.564 & 0.919 & 0.590 & 0.931 & 25 \\
            \hline
            ReSeND                  & \textbf{0.666} & \underline{0.912} & \underline{0.572} & 0.934 & \underline{0.727} & 0.878 & 0.566 & 0.942 & \textbf{0.705} & \underline{0.860} & \textbf{0.659} & \underline{0.911} & \textbf{0.649} & 0.906 & 25 \\
            ReSeND-H               & 0.639 & 0.938 & \textbf{0.583} & \textbf{0.919} & 0.720 & 0.864 & \textbf{0.590} & \textbf{0.899} & \underline{0.679} & 0.895 & \underline{0.637} & 0.914 & 0.641 & \underline{0.905} & 25 \\
            \hline
            \textit{k}-NN (k=1)              & \underline{0.662} & \textbf{0.902} & 0.560 & 0.934 & \textbf{0.754} & \textbf{0.781} & \underline{0.584} & \underline{0.908} & 0.666 & \textbf{0.836} & 0.627 & \textbf{0.900} & \underline{0.642} & \textbf{0.877} & 4800 \\
            \hline
        \end{tabular}
    \end{adjustbox}
    \label{tab:baselines_cross}\vspace{-3mm}
\end{table}

%% file: content/sections/5-conclusions.tex
\section{Conclusions}
In this work, we focused on the OOD detection task considering methods that do not need a fine-tuning stage on the ID data in order to detect semantic novelties. We analyzed how different learning objectives influence the performance of a relational reasoning-based solution to this problem, showing that a lower inter-class separation leads to better generalization. Exploiting this finding we proposed to use a tailored hinge loss function that provides better results than the original method implementation. At the same time, we pointed out how a previously unexplored fine-tuning-free \textit{k}-NN strategy for OOD detection provides unexpectedly good accuracy at the cost of a higher computational effort. Still, it may fail when the support and the test set are drawn from different visual domains.

\noindent \textbf{Acknowledgments} This study was carried out within the FAIR - Future Artificial Intelligence Research and received funding from the European Union Next-GenerationEU (PIANO NAZIONALE DI RIPRESA E RESILIENZA (PNRR) – MISSIONE 4 COMPONENTE 2, INVESTIMENTO 1.3 – D.D. 1555 11/10/2022, PE00000013). This manuscript reflects only the authors’ views and opinions, neither the European Union nor the European Commission can be considered responsible for them.

\noindent Computational resources were provided by IIT HPC infrastructure.